%% file: main.tex
\documentclass[10pt,twocolumn,letterpaper]{article}

\usepackage{cvpr}
\usepackage{times}

\usepackage{amsmath,amssymb,amsfonts,amsthm, dsfont}
\usepackage[usenames,dvipsnames]{color}       % color
\usepackage{algorithm,algorithmicx,listings}      % algorithms
\usepackage{algpseudocode}        % necessary for algorithmicx
\usepackage{epsfig}
\usepackage{graphicx}
\usepackage{tabularx}
\usepackage{makecell}
\usepackage{caption}
\usepackage{subcaption}
\usepackage[breaklinks=true,bookmarks=false,letterpaper=true,%
            colorlinks,citecolor=Black, urlcolor=Violet]{hyperref}

\def\argmin{\mathop{\arg\min}\limits} %
\def\argmax{\mathop{\arg\max}\limits} %

\newcommand{\indicator}{\mathds{1}}
\def\negl{\scalebox{0.75}{$\boldsymbol{\ominus}$}}
\def\posl{\scalebox{0.75}{$\boldsymbol{\oplus}$}}

\newtheorem*{assumption*}{Assumption}
\newtheorem*{problem*}{Problem}

\cvprfinalcopy % *** Uncomment this line for the final submission

\def\cvprPaperID{****} % *** Enter the CVPR Paper ID here

\begin{document}

%%%%%%%%% TITLE
\title{Active Deformable Part Models\\ \normalsize{Technical Report}}

\author{Menglong Zhu \; Nikolay Atanasov \; 
George J. Pappas \; Kostas Daniilidis\\
GRASP Laboratory, University of Pennsylvania\\
3330 Walnut Street, Philadelphia, PA 19104, USA% <-this % stops a space
\thanks{Financial support through the following grants: NSF-IIP-0742304, NSF-OIA-1028009, ARL MAST CTA W911NF-08-2-0004, ARL Robotics CTA W911NF-10-2-0016, NSF-DGE-0966142, NSF-IIS-1317788 and TerraSwarm, one of six centers of STARnet, a Semiconductor Research Corporation program sponsored by MARCO and DARPA is gratefully acknowledged.}\\
{\tt\small \{menglong,atanasov,pappasg,kostas\}@seas.upenn.edu}
% For a paper whose authors are all at the same institution,
% omit the following lines up until the closing ``}''.
% Additional authors and addresses can be added with ``\and'',
% just like the second author.
% To save space, use either the email address or home page, not both
% \and
% Second Author\\
% Institution2\\
% First line of institution2 address\\
% {\tt\small secondauthor@i2.org}
 }

\maketitle
\begin{abstract}
This paper presents an active approach for part-based object detection, which optimizes the order of part filter evaluations and the time at which to stop and make a prediction. Statistics, describing the part responses, are learned from training data and are used to formalize the part scheduling problem as an \textit{offline} optimization. Dynamic programming is applied to obtain a policy, which balances the number of part evaluations with the classification accuracy. During inference, the policy is used as a look-up table to choose the \textit{part order} and the \textit{stopping time} based on the observed filter responses. The method is faster than cascade detection with deformable part models (which does not optimize the part order) with negligible loss in accuracy when evaluated on the PASCAL VOC 2007 and 2010 datasets.
\end{abstract}

%===========================================================
%=== Sections
\input{tex/Introduction}

\input{tex/RelatedWork}

\input{tex/Solution}

\input{tex/Experiments}

\input{tex/Conclusion}

{\small
\bibliographystyle{ieee}
\bibliography{bib/bibref_definitions_short,bib/ref}
}

\end{document}

%% file: tex/Introduction.tex
\section{Introduction}
\label{sec:intro}
Part-based models such as deformable part models (DPM) \cite{felzenszwalb2010object} have become the state of the art in today's object detection methods. They offer powerful representations which can be learned from annotated datasets and capture both the appearance and the configuration of the parts. DPM-based detectors achieve unrivaled accuracy on standard datasets but their computational demand is high since it is proportional to the number of parts in the model and the number of locations at which to evaluate the part filters. Approaches for speeding-up the DPM inference such as cascades, branch-and-bound, and multi-resolution schemes, use the responses obtained from initial part-location evaluations to reduce the future computation. This paper introduces two novel ideas, which are missing in the state-of-the-art methods for speeding up DPM inference.

\begin{figure*}[t!]
    \centering
    \includegraphics[width=\textwidth]{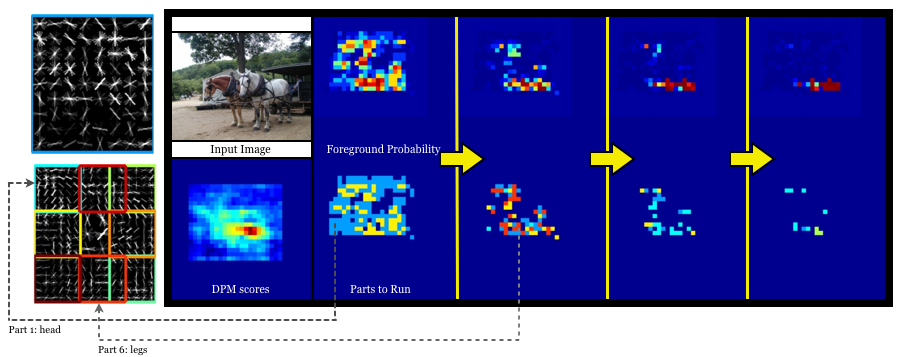}
    \caption{\textbf{Active DPM Overview}: A deformable part model trained on the PASCAL VOC 2007 horse class is shown with colored root and parts in the first column. The second column contains an input image and the original DPM scores as a baseline. The rest of the columns illustrate the inference process of the Active DPM, which proceeds in rounds. The foreground probability (of a horse being present) is maintained at each image location (top row) and is updated sequentially based on the responses of the part filters (high values are red; low values are blue). A policy (learned off-line) is used to select the best sequence of parts to apply at different locations. The bottom row shows the part filters applied at consecutive rounds with colors corresponding to the parts on the left. The policy decides to stop the inference at each location based on the confidence of foreground. As a result, the complete sequence of part filters is evaluated at very few locations, leading to a significant speed-up versus the traditional DPM inference. Our experiments show that the accuracy remains unaffected.}
    \label{fig:big_picture}
\end{figure*}

First, at each location in the image pyramid, a part-based detector has to make a decision: whether to evaluate more parts and in what order or to stop and predict a label. This decision can be regarded as a \textit{planning problem}, whose state space consists of the set of previously used parts and the confidence of whether an object is present or not. While existing approaches rely on a predetermined sequence of parts, our approach optimizes the order in which to apply the part filters so that a minimal number of part evaluations provides maximal classification accuracy at each location. Our second idea is to use a decision loss in the optimization, which quantifies the trade-off between false positive and false negative mistakes, instead of the threshold-based stopping criterion utilized by most other approaches. These ideas have enabled us to propose a novel object detector, Active Deformable Part Models, named so because of the active part selection. The detection procedure consists of two phases: an off-line phase, which learns a part scheduling policy from the training data and an online phase (inference), which uses the policy to optimize the detection task on test images. During inference, each image location starts with equal probabilities for object and background. The probabilities are updated sequentially based on the responses of the part filters suggested by the policy. At any time, depending on the probabilities, the policy might terminate predicting either a background label (which is what most cascaded methods take advantage of) or a positive label, in which case all unused part filters are evaluated in order to obtain the complete DPM score. Fig. \ref{fig:big_picture} exemplifies the inference process.

% , the PASCAL VOC 2010, and the INRIA Person \cite{Dalal_CVPR05}
We evaluated our approach on the PASCAL VOC 2007 and 2010 datasets \cite{Everingham10} and achieved state of the art accuracy but with a 7 times reduction in the number of part-location evaluations and an average speed-up of 3 times compared to the cascade DPM \cite{felzenszwalb2010cascade}. This paper makes the following \textbf{contributions} to the state of the art in part-based object detection:
\begin{enumerate}
  \item We obtain an active part selection policy which optimizes the order of the filter evaluations and balances number of evaluations used with the classification accuracy based on the scores obtained during inference.
  \item The proposed detector achieves a significant speed-up versus the cascade DPM without sacrificing accuracy.
  \item The approach can be generalized to any detection problem, which involves a linear additive score and uses several parts (stages) even if they are just SIFT points.
\end{enumerate}

%% file: tex/RelatedWork.tex
\section{Related Work}
\label{sec:rel_work}
We will refer to work on object detection that optimizes the inference stage rather than the representations since our representation is identical with DPM \cite{felzenszwalb2010object}. 
Our method is inspired by an acceleration of the DPM object detector, the cascade DPM \cite{felzenszwalb2010cascade}. However, while the sequence of parts evaluated in the cascade DPM is pre-defined and a set of thresholds has to be determined empirically, our approach selects the part order and the stopping time at each location based on an optimization criterion. We find the next closest approach to be \cite{Sznitman_2013_CVPR}, which maintains a foreground probability at each stage of a multi-stage ensemble classifier and determines a stopping time based on the corresponding entropy. The difference of our approach is that it jointly optimizes the stage order and the stopping criterion. Kokkinos \cite{kokkinos2011rapid} uses Branch-and-Bound (BB) to prioritize the search over image locations driven by an upper bound on the classification score. It is related to our approach in that object-less locations are easily detected and the search is guided in location space but with the difference that our policy proposes the next part to be tested in cases that no label can yet be given to a particular location. Earlier approaches \cite{lampert08cvpr,lehmann2011fast,lampert2010efficient} relied on BB to constrain the search space of object detectors based on a sliding window or a Hough transform but without deformable parts. Another related group of approaches focuses on learning a sequence of object template tests in position, scale, and orientation space that minimizes the total computation time through a coarse-to-fine evaluation \cite{fleuret2001coarse,pedersoli2011coarse}.

The classic work by Viola and Jones \cite{viola2001rapid} intorduced a cascade of classifiers whose order was determined by importance weights learned by AdaBoost. The approach was studied extensively in \cite{brubaker2008design,zhang2011cvpr,lehmann2011branch,gualdi2012multistage,bourdev2005robust}. Recently, Dollar et al. \cite{dollar2012crosstalk} introduced cross-talk cascades which allow detector responses to trigger or suppress the evaluation of weak classifiers in their neighborhood by exploiting the correlation of the classifier responses in the neighboring positions and scales. Weiss et al. \cite{weiss2012structured} used structured prediction cascades to optimize a function with two objectives: pose refinement and minimum filter evaluation cost. Sapp et al. \cite{sapp2010cascaded} learn a cascade of pictorial structures with increasing pose resolution by progressively filtering the pose state space. Its emphasis is on pre-filtering structures through max-margin scoring rather than part locations so that human poses with weak individual part appearances can still be recovered. Rahtu et al. \cite{rahtu2011learning} use general ``objectness'' filters to produce location proposals which are fed into a cascade, designed to maximize the quality of the locations that advance to the next stage. Our approach is also related to and could be combined with active learning using Gaussian processes for classification \cite{kapoor2010gaussian}.

Similarly to the closest approaches above \cite{felzenszwalb2010cascade,kokkinos2011rapid,Sznitman_2013_CVPR}, our method aims to balance the number of part filter evaluations with the classification accuracy in part-based object detection. The novelty and the main advantage of our approach is that in addition it optimizes the part filter ordering. Since our ``cascades'' still run only on parts, we do not expect the approach to show higher accuracy than structured prediction cascades \cite{sapp2010cascaded} which consider more sophisticated representations that the pictorial structures in the DPM.

%% file: tex/Solution.tex
\section{Technical approach}
\label{sec:tech_app}
The state-of-the-art performance in object detection is obtained by star-structured models such as DPM \cite{felzenszwalb2010object}. A star-structured model of an object with $n$ parts is formally defined by a $(n+2)$-tuple $(F_0,P_1,\ldots,P_n,b)$, where $F_0$ is a root filter, $b$ is a real-valued bias term, and $P_k$ are the part models. Each part model $P_k = (F_k,v_k,d_k)$ consists of a filter $F_k$, a position $v_k$ of the part relative to the root, and the deformation coefficients $d_k$ of a quadratic function specifying a deformation cost for placing the part away from $v_k$.

The object detector is applied in a sliding window fashion and outputs a prediction, $\textit{score}(x)$, at each location $x$ in an image pyramid, where $x = (r,c,l)$ specifies a position $(r,c)$ in the $l$-th level (scale) of the pyramid. The space of all possible locations (position-scale tuples) in the image pyramid is denoted by $\mathcal{X}$. The response of the detector at a given root location $x = (r,c,l) \in \mathcal{X}$ is:
\begin{align*}
    \textit{score}(x) &= F_0'\cdot \phi(H,x) \\ &+ \sum_{k=1}^n \max_{x_k} \biggl(F_k' \cdot \phi(H,x_k) - d_k \cdot \phi_d(\delta_k) \biggr) + b,
\end{align*}
where $\phi(H,x)$ is the histogram of oriented gradients (HOG) feature vector at location $x$ and $\delta_k := (r_k,c_k) - (2(r,c)+v_k)$ is the displacement of the $k$-th part from its anchor position $v_k$ relative to the root location $x$. Each term in the above sum implicitly depends on $x$ since the part locations $x_k$ are chosen relative to root location at $x$. The score can be written as:
\begin{equation}
\label{eq:dpm_score}
\textstyle{
\textit{score}(x) = \sum_{k=0}^n m_k(x)} + b,
\end{equation}
where $m_0(x) := F_0'\cdot \phi(H,x)$ and for $k > 0$, $m_k(x) := \max_{x_k} \bigl(F_k' \cdot \phi(H,x_k) - d_k \cdot \phi_d(\delta_k) \bigr)$. From this perspective, there is no difference between the root and the parts and we can think of the model as one consiting of $n+1$ parts.

%In view of (\ref{eq:dpm_score}), the DPM response can be considered as the sum of the responses of $n+1$ linear filters. 

\subsection{Score Likelihoods for the Parts}
\label{subsec:score_like}
The object detection task requires labeling every $x \in \mathcal{X}$ with a label $y(x) \in \{\negl, \posl\}$. The traditional approach is to compute the complete score in (\ref{eq:dpm_score}) at every position-scale tuple $x \in \mathcal{X}$. In this paper, we argue that it is not necessary to obtain all $n+1$ part responses in order to label a location $x$ correctly. Treating the part scores as noisy observations of the true label $y(x)$, we choose an effective order in which to receive observations and an optimal time to stop. The stopping criterion is based on a trade-off between the cost of obtaining more observations and the cost of labeling the location $x$ incorrectly.

Formally, the part scores $m_0,\ldots,m_n$ at a fixed location $x$ are random variables, which depend on the input image, i.e. the true label $y(x)$. To emphasize this we denote them with upper-case letters $M_k$ and their realizations with lower-case letters $m_k$. In order to predict an effective part order and stopping time, we need statistics which describe the part responses. Let $h^{\posl}(m_0, m_1, \dots, m_n)$ and $h^{\negl}(m_0, m_1, \dots, m_n)$ denote the joint probability density functions (pdf) of the part scores conditioned on the true label being positive $y = \posl$ and negative $y = \negl$, respectively. We make the following assumption.

\begin{assumption*}
The responses of the parts of a star-structured model with a given root location $x \in \mathcal{X}$ are independent conditioned on the the true label $y(x)$, i.e.
\begin{align}
    h^{\posl}(m_0, m_1, \dots, m_n ) &= \textstyle{\prod_{k=0}^n  h_k^{\posl}( m_k )}, \label{eq:cond_indep}\\
    h^{\negl}(m_0, m_1, \dots, m_n ) &= \textstyle{\prod_{k=0}^n  h_k^{\negl}( m_k )}, \notag
\end{align}
where $h_k^{\posl}(m_k)$ is the pdf of $M_k \mid y = \posl$ and $h_k^{\negl}(m_k)$ is the pdf of $M_k \mid y = \negl$.
\end{assumption*}

We learn non-parametric representations for the $2(n+1)$ pdfs $\{h_k^{\posl}, h_k^{\negl}\}$ from an annotated set $D$ of training images. We emphasize that the above assumption does not always hold in practice but simplifies the representation of the score likelihoods significantly\footnote{Removing the independence assumption would require learning the 2 joint $(n+1)$ dimensional pdfs of the part scores in (\ref{eq:cond_indep}) and extracting the $2(n+1)$ marginals and the $2(n+1)(2^n-1)$ conditionals of the form $h( m_k \mid m_I)$, where $I \subseteq \{0,\ldots,n\} \setminus \{k\}$.}. Our algorithm for choosing a part order and a stopping time can be used without the independence assumption. However, we expect the performance to be similar while an unreasonable amount of training data would be required to learn a good representation of the joint pdfs. To evaluate the fidelity of the decoupled representation in (\ref{eq:cond_indep}) we computed correlation coefficients between all pairs of part responses (Table \ref{table:partcorr}) for the classes in the PASCAL 2007 dataset. The mean over all classes, 0.23, indicates a weak correlation. We observed that the few highly correlated parts have identical appearances (e.g. car wheels) or a spatial overlap.

\begin{table*}[t!]
\centering
    \renewcommand{\arraystretch}{1.3}% Wider
    \resizebox{\textwidth}{!}{%
    \begin{tabular}{l*{20}{c}}
        \Xhline{4\arrayrulewidth}
         aero & bike & bird & boat & bottle & bus  & car & cat & chair & cow & table & dog & horse & mbike & person & plant & sheep & sofa & train & tv & \textbf{mean} \\
        \Xhline{4\arrayrulewidth}
        0.36 & 0.37 & 0.14 & 0.18 & 0.24 & 0.29 & 0.40 & 0.16 & 0.13 & 0.17 & 0.44 & 0.11 & 0.23 & 0.21 & 0.14 & 0.21 & 0.26 & 0.22 & 0.24 & 0.20 & 0.23 \\
        \Xhline{4\arrayrulewidth}
    \end{tabular}%
    }
    \caption{Average correlation coefficients among pairs of part responses for all $20$ classes in the PASCAL 2007 dataset}
    \label{table:partcorr}
\end{table*}

\begin{figure}[t!]
    \centering
    \includegraphics[width=\columnwidth]{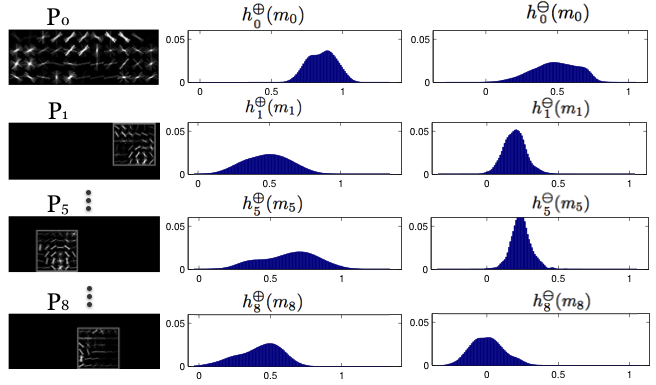}
    \caption{Score likelihoods for several parts from a car DPM model. The root ($P_0$) and three parts of the model are shown on the left. The corresponding positive and negative score likelihoods are shown on the right.}
    \label{fig:likelihood}
\end{figure}

To learn representations for the score likelihoods, $\{h_k^{\posl},h_k^{\negl}\}$, we collected a set of scores for each part from the the training set $D$. Given a positive example $I_i^{\posl} \in D$ of a particular DPM component, the root was placed at the scale and position $x^*$ of the top score within the ground-truth bounding box. The response $m_0^i$ of the root filter was recorded. The parts were placed at their optimal locations relative to the root location $x^*$ and their scores $m_k^i, \; k >0$ were recorded as well. This procedure was repeated for all positive examples in $D$ to obtain a set of scores $\{m_k^i \mid \posl\}$ for each part $k$. For negative examples, $x^*$ was selected randomly over all locations in the image pyramid and the same procedure was used to obtain the set $\{m_k^i \mid \negl\}$. Kernel density estimation was applied to the score collections in order to obtain smooth approximations to $h_k^{\posl}$ and $h_k^{\negl}$. Fig. \ref{fig:likelihood} shows several examples of the score likelihoods obtained from the part responses of a car model.

\subsection{Active Part Selection}
\label{sec:act_learn}
This section discusses how to select an ordered subset of the $n+1$ parts, which when applied at a given location $x \in \mathcal{X}$ has a small probability of mislabeling $x$. The detection at $x$ proceeds in rounds $t = 0,\ldots,n+1$. The DPM inference applies the root and parts in a predefined topological ordering of the model structure. Here, we do not fix the order of the parts a priori. Instead, we select which part to run next \textit{sequentially}, depending on the part responses obtained in the past. The part chosen at round $t$ is denoted by $k(t)$ and can be any of the parts that have not been applied yet. We take a Bayesian approach and maintain a probability $p_t := \mathbb{P}(y = \posl \mid m_{k(0)},\ldots,m_{k(t-1)})$ of a positive label at location $x$ conditioned on the part scores from the previous rounds. The state at time $t$ consists of a binary vector $s_t \in \{0,1\}^{n+1}$ indicating which parts have already been used and the information state $p_t \in [0,1]$. Let $S_t := \{ s \in \{0,1\}^{n+1} \mid \mathbf{1}^Ts = t\}$ be the set\footnote{\textit{Notation}: $\mathbf{1}$ denotes a vector with all elements equal to one, $\mathbf{0}$ denotes a vector with all elements equal to zero, and $e_i$ denotes a vector with one in the $i$-th element and zero everywhere else.} of possible values for $s_t$. At the start of a detection, $s_0 = \mathbf{0}$ and $p_0 = 1/2$, since no parts have been used and we have an uninformative prior for the true label.

Suppose that part $k(t)$ is applied at time $t$ and its score is $m_{k(t)}$. The indicator vector $s_t$ of used parts is updated as:
\begin{equation}
\label{eq:s_dynamics}
s_{t+1} = s_t + e_{k(t)}.
\end{equation}
Due to the independence of the score likelihoods (\ref{eq:cond_indep}), the posterior label distribution is computed using Bayes rule:
\begin{equation}
\label{eq:p_dynamics}
p_{t+1} = \frac{h_{k(t)}^{\posl}(m_{k(t)})}{h_{k(t)}^{\posl}(m_{k(t)}) + h_{k(t)}^{\negl}(m_{k(t)})} p_t.
\end{equation}
In this setting, we seek a conditional plan $\pi$, which chooses which part to run next or stops and decides on a label for $x$. Formally, such a plan is called a \textit{policy} and is a function $\pi(s,p) : \{0,1\}^{n+1} \times [0,1] \rightarrow \{\negl,\posl,0,\ldots,n\}$, which depends on the previously used parts $s$ and the label distribution $p$. An admissible policy does not allow part repetitions and satisfies $\pi(\mathbf{1},p) \in \{\negl,\posl\}$ for all $p \in [0,1]$, i.e. has to choose a label after all parts have been used. The set of admissible policies is denoted by $\Pi$.

Let $\tau(\pi) := \inf \{ t \geq 0 \mid \pi(s_t,p_t) \in \{\negl,\posl\}\} \leq n+1$ denote the stopping time of policy $\pi \in \Pi$. Let $\hat{y}_\pi \in \{\negl,\posl\}$ denote the label guessed by policy $\pi$ after its termination. We would like to choose a policy, which decides \textit{quickly} and \textit{correctly}. To formalize this, define the probability of making an error as $Pe(\pi) := \mathbb{P}(\hat{y}_\pi \neq y)$, where $y$ is the hidden correct label of $x$.

\begin{problem*}[Active Part Selection]
Given $\epsilon > 0$, choose an admissible part policy $\pi$ with minimum expected stopping time and probability of error bounded by $\epsilon$:  
\begin{align}
\min_{\pi \in \Pi} \quad &\mathbb{E}[ \tau(\pi) ] \label{eq:act_part_sel}\\
\text{s.t.} \quad &Pe(\pi) \leq \epsilon, \notag
\end{align}
where the expectation is over the hidden label $y$ and the part scores $M_{k(0)}, \ldots, M_{k(\tau-1)}$.
\end{problem*}

Note that if $\epsilon$ is chosen too small, (\ref{eq:act_part_sel}) might be infeasible. In other words, even the best sequencing of the parts might not reduce the probability of error sufficiently. To avoid this issue, we relax the constraint in (\ref{eq:act_part_sel}) by introducing a Lagrange multiplier $\lambda > 0$ as follows:
\begin{equation}
\label{eq:lag_relax}
\min_{\pi \in \Pi} \quad \mathbb{E}[ \tau(\pi) ] + \lambda Pe(\pi).
\end{equation}
The Lagrange multiplier $\lambda$ can be interpreted as a cost paid for choosing an incorrect label. To elaborate on this, we rewrite the cost function as follows:
\begin{align*}
\mathbb{E} \biggl[ \tau + \lambda \mathbb{E}_y \bigl[  &\indicator_{\{\hat{y} \neq y\}} \mid M_{k(0)},\ldots,M_{k(\tau-1)} \bigr] \biggr] \\
= \mathbb{E}\biggl[ \tau &+ \lambda \indicator_{\{\hat{y} \neq \posl\}} \mathbb{P}\bigl( y = \posl \mid M_{k(0)},\ldots,M_{k(\tau-1)} \bigr) \\
 &+ \lambda \indicator_{\{\hat{y} \neq \negl\}} \mathbb{P}\bigl( y = \negl \mid M_{k(0)},\ldots,M_{k(\tau-1)}  \bigr) \biggr] \\
= \mathbb{E} \biggl[ \tau &+ \lambda p_\tau \indicator_{\{\hat{y} = \negl\}}  + \lambda (1-p_\tau) \indicator_{\{\hat{y} = \posl \}} \biggr].
%\longeq{(\ref{eq:decision})}{} \mathbb{E} \biggl( \tau &+ \lambda p_\tau \indicator_{\{p_\tau \leq 1 - p_\tau\}}  + \lambda (1-p_\tau) \indicator_{\{1-p_\tau < p_\tau \}} \biggr).
%\longeq{(\ref{eq:decision})}{} \mathbb{E} \biggl( \tau &+ \lambda \min\{p_\tau, 1-p_\tau\} \biggr)
\end{align*}
The term $\lambda p_\tau$ above is the cost paid if label $\hat{y} = \negl$ is chosen incorrectly. Similarly, $\lambda (1-p_\tau)$ is the cost paid if label $\hat{y} = \posl$ is chosen incorrectly. To allow flexibility, we introduce separate costs $\lambda_{fp}$ and $\lambda_{fn}$ for false positive and false negative mistakes. The final form of the \textbf{Active Part Selection} problem is:
\begin{equation}
\label{prob:main_prob}
\min_{\pi \in \Pi} \quad \mathbb{E} \biggl[ \tau + \lambda_{fn} p_\tau \indicator_{\{\hat{y} = \negl\}}  + \lambda_{fp} (1-p_\tau) \indicator_{\{\hat{y} = \posl\}} \biggr].
\end{equation}

\paragraph{Computing the Part Selection Policy}
Problem (\ref{prob:main_prob}) can be solved using Dynamic Programming \cite{Bertsekas_DP95}. For a fixed policy $\pi \in \Pi$ and a given initial state $s_0 \in \{0,1\}^{n+1}$ and $p_0 \in [0,1]$, the value function:
\[
V_\pi(s_0,p_0) := \mathbb{E} \biggl[ \tau + \lambda_{fn} p_\tau \indicator_{\{\hat{y} = \negl\}}  + \lambda_{fp} (1-p_\tau) \indicator_{\{\hat{y} = \posl\}} \biggr],
\]
is a well-defined quantity. The \textit{optimal} policy $\pi^*$ and the corresponding \textit{optimal} value function are obtained as:
\begin{align*}
V^*(s_0,p_0) &= \min_{\pi \in \Pi} V_\pi(s_0,p_0),\\
\pi^*(s_0,p_0) &= \argmax_{\pi \in \Pi} V_\pi(s_0,p_0).
\end{align*}
To compute $\pi^*$ we proceed backwards in time. Suppose that the policy has not terminated by time $t = n+1$. Since there are no parts left to apply the policy is forced to terminate. Thus, $\tau = n+1$ and $s_{n+1} = \mathbf{1}$ and for all $p \in [0,1]$ the optimal value function becomes:
\begin{align}
V^*(\mathbf{1},p) &= \min_{\hat{y} \in \{\negl,\posl\}} \biggl\{ \lambda_{fn} p \indicator_{\{\hat{y} = \negl\}}  + \lambda_{fp} (1-p) \indicator_{\{\hat{y} = \posl\}} \biggr\} \notag\\
&= \hspace{8pt} \min \{\lambda_{fn} p, \lambda_{fp} (1-p) \}. \label{eq:dp_final}
\end{align}
The intermediate stage values for $t = n, \ldots, 0$, $s_t \in S_t$, and $p_t \in [0,1]$ are:
\begin{align}
V^*(s_t,p_t) =& \min \biggl\{ \lambda_{fn} p_t, \lambda_{fp}(1-p_t), \label{eq:dp_int}\\
1 + \min_{k \in \mathcal{A}(s_t)}& \mathbb{E}_{M_{k}} V^*\biggl(s_t + e_{k}, \frac{h_k^{\posl}(M_{k})p_t}{h_k^{\posl}(M_{k})+h_k^{\negl}(M_{k})}\biggr)  \biggr\}, \notag
\end{align}
where $\mathcal{A}(s) := \{ i \in \{0,\ldots,n\} \mid s_i = 0\}$ is the set of available (unused) parts\footnote{Each score likelihood was discretized using $201$ bins to obtain a histogram. Then, the expectation in (\ref{eq:dp_int}) was computed as a sum over the bins. Alternatively, Monte Carlo integration can be performed by sampling from the Gaussian mixtures directly.}. The optimal policy is readily obtained from the optimal value function. At stage $t$, if the first term in (\ref{eq:dp_int}) is smallest, the policy stops and chooses $\hat{y} = \negl$; if the second term is smallest, the policy stops and chooses $\hat{y} = \posl$; otherwise, the policy chooses to run the part $k$, which minimizes the expectation.

Alg. \ref{alg:aps} summarizes the steps necessary to compute the optimal policy $\pi^*$ using the score likelihoods $\{h_k^{\posl},h_k^{\negl}\}$ from Sec. \ref{subsec:score_like}. The one dimensional space $[0,1]$ of label probabilities $p$ can be discretized into $d$ bins in order to store the function $\pi$ returned by Alg. \ref{alg:aps}. The memory required is $O(d 2^{n+1})$ since the space $\{0,1\}^{n+1}$ of used-part indicator vectors grows exponentially with the number of parts. Nevertheless, in practice the number of parts in a DPM is rarely more than $20$ and Alg. \ref{alg:aps} can be executed.

% algorithm
\begin{algorithm}[htb]
\caption{Active Part Selection}
\begin{algorithmic}[1]
\footnotesize
\State \textbf{Input}: Score likelihoods $\{ h_k^{\negl}, h_k^{\posl} \}_{k=0}^n$ for all parts, false positive cost $\lambda_{fp}$, false negative cost $\lambda_{fn}$
\State \textbf{Output}: Policy $\pi : \{0,1\}^{n+1} \times [0,1] \rightarrow \{\negl,\posl,0,\ldots,n\}$
\State
\State $S_t := \{ s \in \{0,1\}^{n+1} \mid \mathbf{1}^Ts = t\}$
\State $\mathcal{A}(s) := \{ i \in \{0,\ldots,n\} \mid s_i = 0\}$ for $s \in \{0,1\}^{n+1}$
\State
\State $V(\mathbf{1},p) := \min\{ \lambda_{fn}p, \lambda_{fp}(1-p)\}, \quad \forall p \in [0,1]$
\State $\pi(\mathbf{1},p) := \begin{cases}
  \negl, & \lambda_{fn} p \leq \lambda_{fp}(1-p)\\
  \posl, & \text{otherwise}
\end{cases}$
\State
\For{$t = n, n-1,\ldots,0$}
  \For{$s \in S_t$}
    \For{$k \in \mathcal{A}(s)$}
      \State $Q(s, p, k) := \mathbb{E}_{M_{k}} V\biggl(s+e_{k}, \frac{h_k^{\posl}(M_{k})p}{h_k^{\posl}(M_{k})+h_k^{\negl}(M_{k})}\biggr)$
    \EndFor
	  \State $V(s,p) \!:=\! \min \biggl\{ \lambda_{fn} p, \lambda_{fp}(1-p), 1 + \!\!\underset{k \in \mathcal{A}(s)}{\min}\! Q(s,p,k)  \biggr\}$
	  \State $\pi(s,p) := \begin{cases}
	    \negl, & V(s,p) = \lambda_{fn} p,\\
	    \posl, & V(s,p) = \lambda_{fp}(1-p),\\
	    \argmin_{k \in \mathcal{A}(s)} Q(s,p,k), & \text{otherwise} 
	  \end{cases}$
	\EndFor
\EndFor
\State \textbf{return} $\pi$
\end{algorithmic}
\label{alg:aps}
\end{algorithm}

\subsection{Active DPM Inference}
\label{sec:ADPM_inference}
A policy $\pi$ is obtained \textit{offline} using Alg. \ref{alg:aps}. In the online phase, $\pi$ is used to select a sequence of parts to apply at each location $x \in \mathcal{X}$ in the image pyramid. Note that the labeling of each location is treated as an independent problem and proceeds in parallel. Alg. \ref{alg:inference} summarizes the Active DPM inference process.

% algorithm
\begin{algorithm}[htb]
\caption{Active DPM Inference}
\begin{algorithmic}[1]
\footnotesize
\State \textbf{Input}: Image pyramid, model $(F_0,P_1,\ldots,P_n,b)$, score likelihoods $\{ h_k^{\negl}, h_k^{\posl} \}_{k=0}^n$ for all parts, policy $\pi$ 
\State \textbf{Output}: $score(x)$ at all locations $x \in \mathcal{X}$ in the image pyramid
\State
\For{$x \in 1 \ldots |\mathcal{X}|$} \Comment{All image pyramid locations}
    \State $s_0 := \mathbf{0}$; $p_0 = 0.5$; $score(x) := 0$
    \For{$t = 0, 1,\ldots, n$}
        \State $k := \pi(s_t,p_t)$ \Comment{Lookup next best part}
        \If{$k = \posl $} \Comment{Labeled as foreground}
            \For {$i \in \{0,1,\ldots,n\}$}
                \If {$s_t(i) = 0$}
                \State Compute score $m_k(x)$ for part $k$ \Comment{\textit{O}($|\Delta|$)}
                    \State $score(x) := score(x) + m_k(x)$
                \EndIf
            \EndFor
            \State $score(x) := score(x) + b$ \Comment{Add bias to final score}
            \State break;
        \ElsIf{$k = \negl $} \Comment{Labeled as background}
            \State $score(x) := -\infty$
            \State break;
        \Else \Comment{Update probability and score}
            \State Compute score $m_k(x)$ for part $k$ \Comment{\textit{O}($|\Delta|$)}
            \State $score(x) := score(x) + m_k(x)$
            \State $p_{t+1} := \frac{h_k^{\posl}(m_k(x))p_t}{h_k^{\posl}(m_k(x)) + h_k^{\negl}(m_k(x))}$
            \State $s_{t+1} = s_t + e_k$
        \EndIf
    \EndFor
\EndFor
\end{algorithmic}
\label{alg:inference}
\end{algorithm}

At the start of a detection at location $x$, $s_0 = \mathbf{0}$ since no parts have been used and $p_0 = 1/2$ since we have an uninformative label prior (Line 5). At each round $t$, the policy is queried to obtain either the next part to run or a predicted label for $x$ (Line 7). Note that querying the policy is an $O(1)$ operation since it is stored as a lookup table. If the policy terminates and labels $y(x)$ as foreground (Line 8), all unused part filters are applied in order to obtain the final discriminative score in (\ref{eq:dpm_score}). On the other hand, if the policy terminates and labels $y(x)$ as background, no additional part filters are evaluated and the final score is set to $-\infty$ (Line 18). In this case, our algorithm makes computational savings compared to the DPM. The potential speed-up and the effect on accuracy are discussed in the Sec. \ref{sec:experiments}. Finally, if the policy returns a part index $k$, the corresponding score $m_k(x)$ is computed by applying the part filter (Line 21). This operation is $O(|\Delta|)$, where $\Delta$ is the space of possible displacements for part $k$ with respect to the root location $x$. Following the analysis in \cite{felzenszwalb2010cascade}, searching over the possible locations for part $k$ is usually no more expensive than evaluating its linear filter $F_k$ once because the spatial extent of the filter is of similar size as its range of displacement. This is the case because once $F_k$ is applied at some location $x_k$, the resulting response $\Phi_k(x_k) = F_k' \cdot \phi(H,x_k)$ is cached to avoid recomputing it later. The use of a memorized version $\tilde{\Phi}_k(x_k)$ of $\Phi_k(x_k)$ amortizes the complexity of the search over $\Delta$. The score $m_k$ of part $k$ is used to update the total score at $x$ (Line 22). Then, the dynamics in (\ref{eq:s_dynamics}) and (\ref{eq:p_dynamics}) are used to update the state $(s_t, p_t)$ (Line 23 - 24). Since the policy lookups and the state updates are all of $O(1)$ complexity, the worst-case complexity of Alg. \ref{alg:inference} is $O(n |\mathcal{X}| |\Delta|)$. The worst-case complexity is the same as that of the DPM and the cascade DPM. The average running time of our algorithm depends on the total number of score $m_k$ evaluations, which in turn depends on the choice of the parameters $\lambda_{fn}$ and $\lambda_{fp}$ and is the subject of the next section.

%% file: tex/Experiments.tex
\section{Experiments}
\label{sec:experiments}
% Speed-up vs Accuracy Trade-off
\subsection{Speed-Accuracy Trade-Off} 
\label{sec:speed_acc}
The two parameters of the Active DPM (ADPM) method are the penalty, $\lambda_{fp}$, for incorrectly predicting background as foreground and the penalty, $\lambda_{fn}$, for incorrectly predicting foreground as background. The accuracy and the speed of the ADPM inference depend on these parameters. To get an intuition, consider making both $\lambda_{fp}$ and $\lambda_{fn}$ very small. The cost of an incorrect prediction will be negligible, thus encouraging the policy to sacrifice accuracy and stop immediately. In the other extreme, when both parameters are very large, the policy will delay the prediction as much as possible in order to obtain more information.

\begin{figure}[!t]

    \centering
%\begin{figure}
    %\centering
    %\begin{subfigure}[b]{0.245\columnwidth}
        \includegraphics[width=0.48\columnwidth]{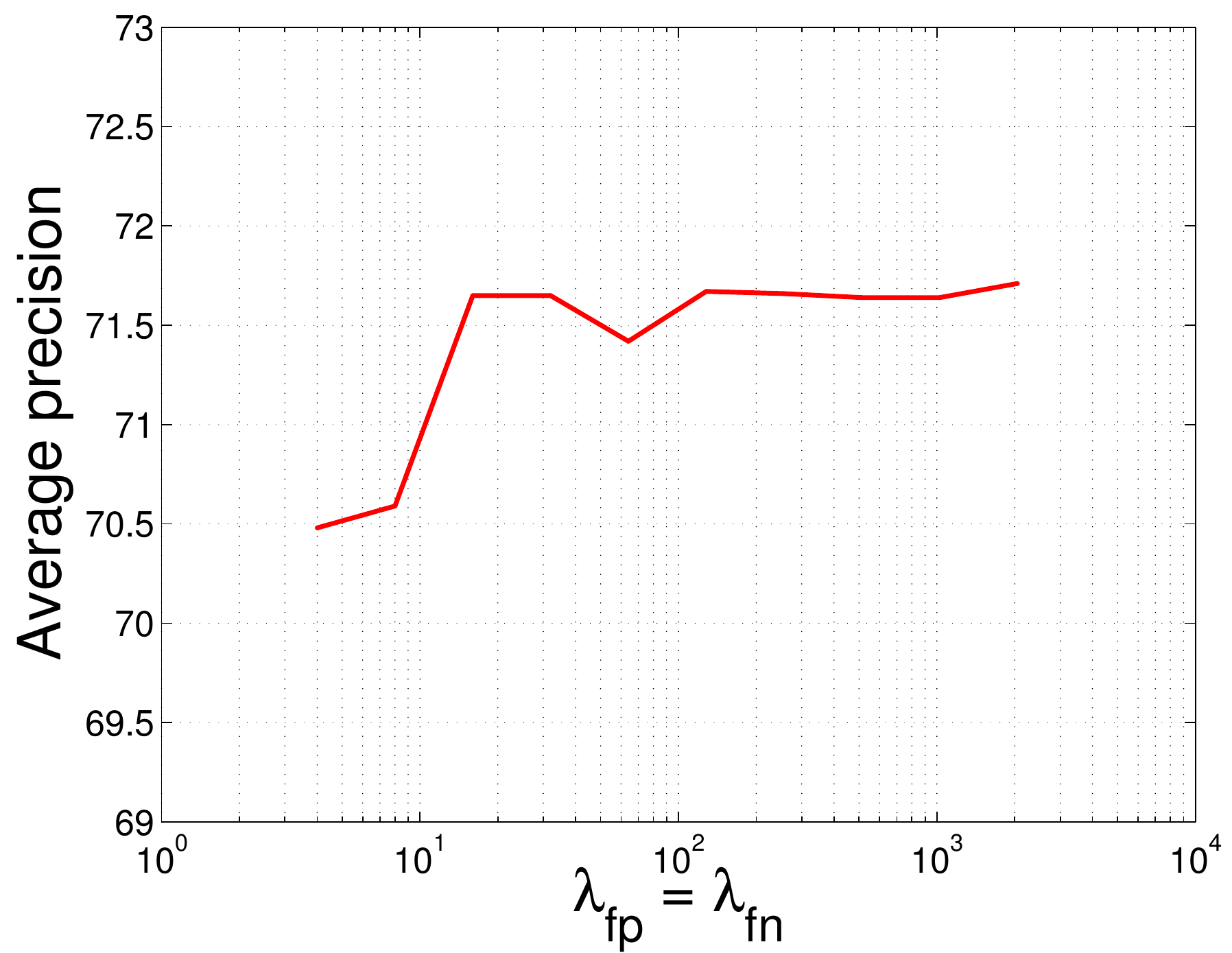}
    %    \label{fig:ap_bus}
    %\end{subfigure}%
    %\begin{subfigure}[b]{0.245\columnwidth}
        \includegraphics[width=0.48\columnwidth]{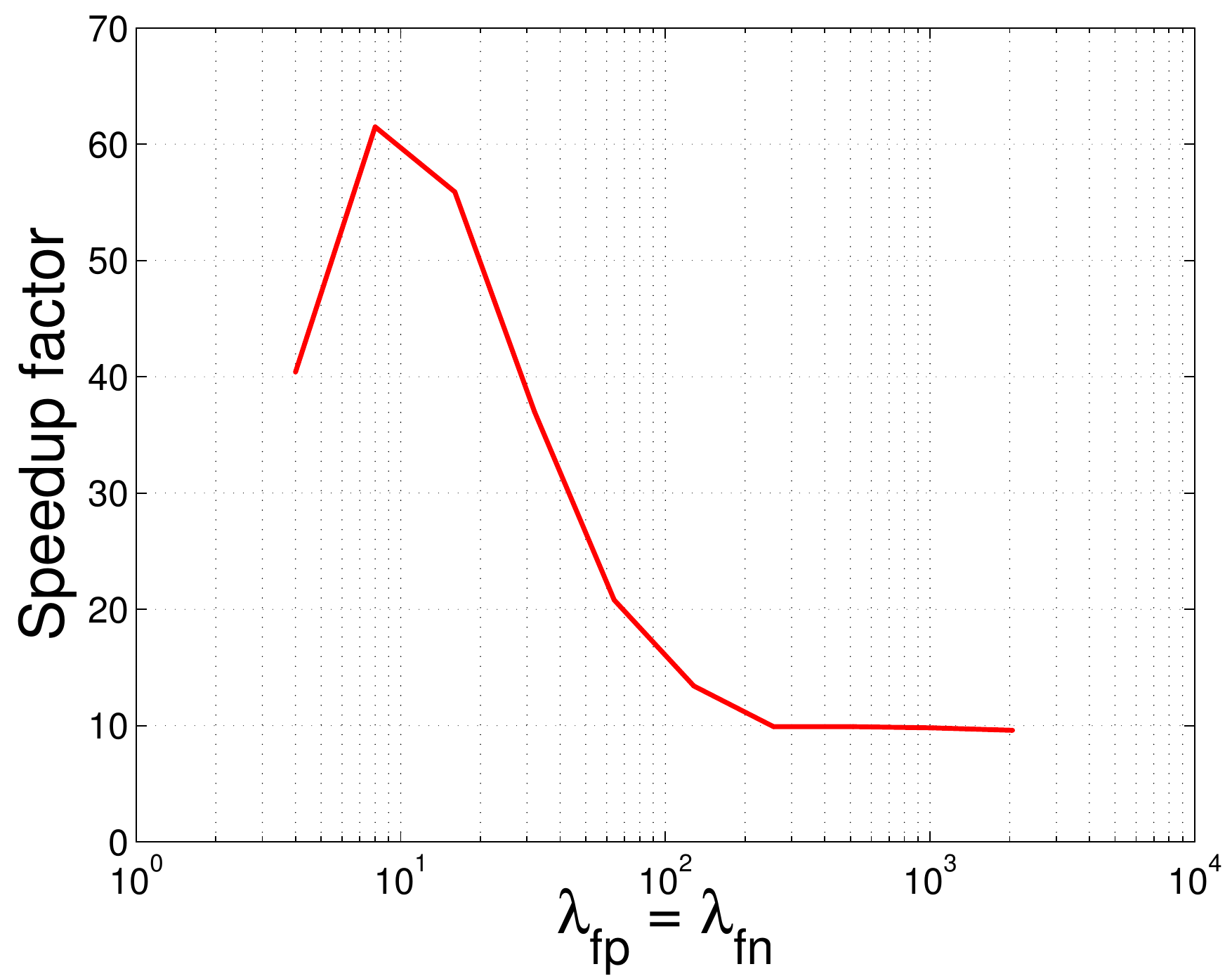}
    %    \label{fig:speedup_bus}
    %\end{subfigure}%
    \captionof{figure}{Average precision and relative number of part evaluations versus DPM as a function of the parameter $\lambda = \lambda_{fn} = \lambda_{fp}$ on a log scale. The curves are reported on the bus class from the VOC 2007 training set.}
    \label{fig:speedup_acc}
\end{figure}

To evaluate the effect of the parameters, we compared the average precision (AP) and the number of part evaluations of Alg. \ref{alg:inference} to those of the traditional DPM as a baseline. Let $R_M$ be the total number of score $m_k(x)$ evaluations for $k > 0$ (excluding the root) over all locations $x \in \mathcal{X}$ performed by method M. For example, $R_{DPM} = n|\mathcal{X}|$ since the DPM evaluates all parts at all locations in $\mathcal{X}$. We define the \textbf{relative number of part evaluations} (RNPE) of our method (ADPM) versus method M as the ratio of $R_M$ to $R_{ADPM}$. The AP and the RNPE versus DPM of APDM were evaluated on several classes from the PASCAL VOC 2007 training set. Fig. \ref{fig:speedup_acc} shows the performance as the parameter $\lambda = \lambda_{fn} = \lambda_{fp}$ is varied. As expected, the AP increases while the RNPE decreases as the penalty of an incorrect declaration $\lambda$ grows because ADPM evaluates more parts. The dip in RNPE for very low $\lambda$ values is due to fact that ADPM starts reporting too many false-positives. In the case of a positive declaration all part responses need to be computed, which reduces the speed-up versus DPM.
%{\color{red}The dip in RNPE for very low $\lambda$ values is due to increasing false negatives as ADPM becomes more aggressive on predicting backgrounds. In the case of a positive declaration all part responses need to be computed, which reduces the RNPE versus DPM.}

Since a positive declaration always requires $n+1$ part evaluations, we limit the number of false positive mistakes made by the policy by setting $\lambda_{fp} > \lambda_{fn}$. While this might hurt the accuracy, it will certainly result in significantly less part evaluations. To verify this intuition we performed experiments with $\lambda_{fp} > \lambda_{fn}$ on the PASCAL VOC 2007 dataset. Table \ref{table:speedup_acc} reports the AP and the RNPE versus DPM from a grid search over the parameter space. Generally, as the ratio between $\lambda_{fp}$ and $\lambda_{fn}$ increases, the RNPE increases while the AP decreases. Notice, however, that the increase in RNPE is significant, while the hit in accuracy is negligible.

% Similar trends were observed on the other classes.

\begin{table}[!t]
    \centering
    %\vspace{-5pt}
%    \renewcommand{\arraystretch}{1}% Wider
    \resizebox{1\columnwidth}{!}{%
    \begin{tabular}{*{12}{|c|}}
        \Xhline{4\arrayrulewidth}
            \multicolumn{6}{|c||}{Average Precision} & \multicolumn{6}{c|}{RNPE vs DPM} \\
            \hline
            $\lambda_{fp}/\lambda_{fn}$ & 4 & 8 & 16 & 32 & 64 & $\lambda_{fp}/\lambda_{fn}$ & 4 & 8 & 16 & 32 & 64 \\
            \Xhline{4\arrayrulewidth}
            4 & 70.3 & & & & & 4 & 40.4 & & & & \\
            \hline
            8 & 70.0  & 71.0 & & & & 8 & 80.7 & 61.5 & & & \\
            \hline
            16 & 69.6 & 71.1 & 71.5 & & & 16 & 118.6 & 74.5 & 55.9 & & \\
            \hline
            32 & 70.5 & 70.7  & 71.6 & 71.6 & & 32 & 178.3 & 82.1 & 59.8 & 37.0 & \\
            \hline
            64 & 67.3 & 69.6 &  71.5 & 71.6 & 71.4 & 64 & 186.9 & 96.4 & 56.2 & 34.5 & 20.8 \\
        \Xhline{4\arrayrulewidth}
    \end{tabular}
    }
    %\vspace{6pt}
    \captionof{table}{Average precision and relative number of part evaluations versus DPM obtained on the bus class from PASCAL VOC 2007 training set. A grid search over the parameter space $(\lambda_{fp},\lambda_{fn}) \in \{4,8,\ldots,64\}\times \{4,8,\ldots,64\}$ with $\lambda_{fp} \geq \lambda_{fn}$ is shown.}
    \label{table:speedup_acc}
\end{table}

%& INRIA
%& 65.3
%& 2.7 
%& 88.0
%& 88.0 
%& 87.7
%+=========================== RESULTS ================================%

\subsection{Results}
\label{sec:results}

\begin{table*}[t]
% Speed-up table
    \centering
    \renewcommand{\arraystretch}{1.5}% Wider
    \resizebox{\textwidth}{!}{%
    \begin{tabular}{r*{21}{c}}
        \Xhline{5\arrayrulewidth}
        \textbf{VOC2007} & aero & bike & bird & boat & bottle & bus  & car & cat & chair & cow & table & dog & horse & mbike & person & plant & sheep & sofa & train & tv & \textbf{mean}   \\
        \Xhline{3\arrayrulewidth}
         DPM RNPE   & 102.8 & 106.7 & 63.7 & 79.7 & 58.1 & 155.2 & 44.5 & 40.0 & 58.9 & 71.8 & 69.9 & 49.2 & 51.0 & 59.6 & 45.3 & 49.0 & 62.6 & 68.6 & 79.0 & 100.6 & \textbf{70.8}  \\
        %\#RNPE Cascade   & 30.2 & 8 & 29.3 & 31.7 & 26.0 & 15.2 & 11.6 & 8.8 & 25.3 & 11.1 & 8.2 & 12.7 & 5.1 & 6.6 & 15.1 & 26.6 & 19.4 & 7.3 & 4.4 & 28.0 & \textbf{16.5}   \\
        \Xhline{3\arrayrulewidth}
        DPM AP            & 33.2 & 60.3 & 10.2 & 16.1 & 27.3 & 54.3 & 58.2 & 23.0 & 20.0 & 24.1 & 26.7 & 12.7 & 58.1 & 48.2 & 43.2 & 12.0 & 21.1 & 36.1 & 46.0 & 43.5 & \textbf{33.7}  \\
        %Cascade AP       & 33.2 & 60.8 & 10.2 & 16.1 & 27.3 & 54.1 & 58.1 & 23.0 & 20.0 & 24.2 & 26.8 & 12.7 & 58.1 & 48.2 & 43.2 & 12.0 & 20.1 & 35.8 & 46.0 & 43.4 & \textbf{33.7}  \\
        ADPM AP    & 33.5 & 59.8 & 9.8 & 15.3 & 27.6 & 52.5 & 57.6 & 22.1 & 20.1 & 24.6 & 24.9 & 12.3 & 57.6 & 48.4 & 42.8 & 12.0 & 20.4 & 35.7 & 46.3 & 43.2 & \textbf{33.3}  \\
        
        \Xhline{5\arrayrulewidth}

        \textbf{VOC2010} & aero & bike & bird & boat & bottle & bus  & car & cat & chair & cow & table & dog & horse & mbike & person & plant & sheep & sofa & train & tv & \textbf{mean} \\
        \Xhline{3\arrayrulewidth}
        %\#RNPE DPM   & 51.5 & 75.3 & 27.3 & 69.8 & 42.2 & 132.0 & 40.2 & 60.3 & 35.5 & 104.7 & 53.3 & 60.7 & 64.7 & 106.4 & 46.1 & 38.6 & 21.8 & 89.6 & 98.2 & 67.4 & 64.3 \\
        %\#RNPE Cascade   & 5.2 & 3.6 & 8.3 & 21.9 & 11.3 & 9.0 & 7.7 & 9.4 & 9.9 & 10.8 & 5.4 & 9.6 & 5.1 & 11.5 & 8.9 & 14.0 & 3.5 & 11.2 & 5.8 & 10.8 & 9.2 \\
         DPM RNPE  & 110.0 & 100.8 & 47.9 & 98.8 & 111.8 & 214.4 & 75.6 & 202.5 & 150.8 & 147.2 & 62.4 & 126.2 & 133.7 & 187.1 & 114.4 & 59.3 & 24.3 & 131.2 & 143.8 & 106.0 & \textbf{117.4} \\
        %\#RNPE Cascade   & 11.0 & 4.9 & 14.6 & 31.0 & 30.0 & 14.6 & 14.5 & 31.5 & 42.1 & 15.2 & 6.4 & 20.1 & 10.4 & 20.3 & 22.2 & 21.5 & 3.9 & 16.4 & 8.6 & 17.0 & \textbf{17.8} \\
        \Xhline{3\arrayrulewidth}
        DPM AP      & 45.6 & 49.0 & 11.0 & 11.6 & 27.2 & 50.5 & 43.1 & 23.6 & 17.2 & 23.2 & 10.7 & 20.5 & 42.5 & 44.5 & 41.3 & 8.7 & 29.0 & 18.7 & 40.0 & 34.5 & \textbf{29.6} \\
        %Cascade AP  & 45.5 & 48.9 & 11.0 & 11.6 & 27.2 & 50.5 & 43.1 & 23.6 & 17.2 & 23.1 & 10.7 & 20.5 & 42.4 & 44.5 & 41.3 & 8.7 & 29.0 & 18.7 & 40.1 & 34.4 & \textbf{29.6} \\
        ADPM AP     & 45.3 & 49.1 & 10.2 & 12.2 & 26.9 & 50.6 & 41.9 & 22.7 & 16.5 & 22.8 & 10.6 & 19.7 & 40.8 & 44.5 & 36.8 & 8.3 & 29.1 & 18.6 & 39.7 & 34.5 & \textbf{29.1} \\
        % 20_8
        %ADPM AP     & 45.6 & 49.1 & 10.3 & 12.1 & 26.9 & 50.7 & 42.3 & 23.2 & 17.0 & 23.0 & 10.4 & 20.3 & 41.6 & 44.5 & 41.2 & 8.6 & 29.0 & 18.4 & 39.8 & 34.6 & 29.4 \\
        \Xhline{5\arrayrulewidth}
    \end{tabular}
    }
    \caption{Average precision (AP) and relative number of part evaluations (RNPE) of DPM versus ADPM on all $20$ classes in PASCAL VOC 2007 and 2010.}
    \label{table:speedupDPM}
    $ $\\
    \centering
    \renewcommand{\arraystretch}{1.5}% Wider
    \resizebox{\textwidth}{!}{%
    \begin{tabular}{r*{21}{c}}
        \Xhline{5\arrayrulewidth}
        \textbf{VOC2007} & aero & bike & bird & boat & bottle & bus  & car & cat & chair & cow & table & dog & horse & mbike & person & plant & sheep & sofa & train & tv & \textbf{mean}   \\
        \Xhline{3\arrayrulewidth}
        %\#RNPE DPM    & 102.8 & 106.7 & 63.7 & 79.7 & 58.1 & 155.2 & 44.5 & 40.0 & 58.9 & 71.8 & 69.9 & 49.2 & 51.0 & 59.6 & 45.3 & 49.0 & 62.6 & 68.6 & 79.0 & 100.6 & \textbf{70.8}  \\
         Cascade RNPE  & 5.93 & 5.35 & 9.17 & 6.09 & 8.14 & 3.06 & 5.61 & 4.51 & 6.30 & 4.03 & 4.83 & 7.77 & 3.61 & 6.67 & 17.8 & 9.84 & 3.82 & 2.43 & 2.89 & 6.97 & \textbf{6.24}   \\
        ADPM Speedup   & 3.14 & 1.60 & 8.21 & 4.57 & 3.36 & 1.67 & 2.11 & 1.54 & 3.12 & 1.63 & 1.28 & 2.72 & 1.07 & 1.50 & 3.59 & 6.15 & 2.92 & 1.10 & 1.11 & 3.26 & \textbf{2.78}  \\
        \Xhline{3\arrayrulewidth}
        %DPM AP            & 33.2 & 60.3 & 10.2 & 16.1 & 27.3 & 54.3 & 58.2 & 23.0 & 20.0 & 24.1 & 26.7 & 12.7 & 58.1 & 48.2 & 43.2 & 12.0 & 21.1 & 36.1 & 46.0 & 43.5 & \textbf{33.7}  \\
        Cascade AP       & 33.2 & 60.8 & 10.2 & 16.1 & 27.3 & 54.1 & 58.1 & 23.0 & 20.0 & 24.2 & 26.8 & 12.7 & 58.1 & 48.2 & 43.2 & 12.0 & 20.1 & 35.8 & 46.0 & 43.4 & \textbf{33.7}  \\
        ADPM AP    & 31.7 & 59.0 & 9.70 & 14.9 & 27.5 & 51.4 & 56.7 & 22.1 & 20.4 & 24.0 & 24.7 & 12.4 & 57.7 & 48.5 & 41.7 & 11.6 & 20.4 & 35.9 & 45.8 & 42.8 & \textbf{33.0}  \\
        \Xhline{5\arrayrulewidth}

        \textbf{VOC2010} & aero & bike & bird & boat & bottle & bus  & car & cat & chair & cow & table & dog & horse & mbike & person & plant & sheep & sofa & train & tv & \textbf{mean} \\
        \Xhline{3\arrayrulewidth}
        %\#RNPE DPM   & 51.5 & 75.3 & 27.3 & 69.8 & 42.2 & 132.0 & 40.2 & 60.3 & 35.5 & 104.7 & 53.3 & 60.7 & 64.7 & 106.4 & 46.1 & 38.6 & 21.8 & 89.6 & 98.2 & 67.4 & 64.3 \\
        %\#RNPE Cascade   & 5.2 & 3.6 & 8.3 & 21.9 & 11.3 & 9.0 & 7.7 & 9.4 & 9.9 & 10.8 & 5.4 & 9.6 & 5.1 & 11.5 & 8.9 & 14.0 & 3.5 & 11.2 & 5.8 & 10.8 & 9.2 \\
        %\#RNPE DPM   & 110.0 & 100.8 & 47.9 & 98.8 & 111.8 & 214.4 & 75.6 & 202.5 & 150.8 & 147.2 & 62.4 & 126.2 & 133.7 & 187.1 & 114.4 & 59.3 & 24.3 & 131.2 & 143.8 & 106.0 & \textbf{117.4} \\
         Cascade RNPE  & 7.28& 2.66& 14.80& 7.83& 12.22& 5.47& 6.29& 6.33& 9.72& 4.16& 3.74& 10.77& 3.21& 9.68& 21.43& 12.21& 3.23& 4.58& 3.98& 8.17 & \textbf{7.89} \\
         ADPM Speedup & 2.15& 1.28& 7.58& 5.93& 4.68& 2.79& 2.28& 2.44& 3.72& 2.42& 1.52& 2.76& 1.57& 2.93& 4.72& 8.24& 1.42& 1.81& 1.47& 3.41 & \textbf{3.26} \\
        \Xhline{3\arrayrulewidth}
        %DPM AP      & 45.6 & 49.0 & 11.0 & 11.6 & 27.2 & 50.5 & 43.1 & 23.6 & 17.2 & 23.2 & 10.7 & 20.5 & 42.5 & 44.5 & 41.3 & 8.7 & 29.0 & 18.7 & 40.0 & 34.5 & \textbf{29.6} \\
        Cascade AP  & 45.5 & 48.9 & 11.0 & 11.6 & 27.2 & 50.5 & 43.1 & 23.6 & 17.2 & 23.1 & 10.7 & 20.5 & 42.4 & 44.5 & 41.3 & 8.7 & 29.0 & 18.7 & 40.1 & 34.4 & \textbf{29.6} \\
        ADPM AP     & 44.5 & 49.2 & 9.5 & 11.6 & 25.9 & 50.6 & 41.7 & 22.5 & 16.9 & 22.0 & 9.8 & 19.8 & 41.1 & 45.1 & 40.2 & 7.4 & 28.5 & 18.3 & 38.0 & 34.5 & \textbf{28.8} \\
        % 20_8
        %ADPM AP     & 45.6 & 49.1 & 10.3 & 12.1 & 26.9 & 50.7 & 42.3 & 23.2 & 17.0 & 23.0 & 10.4 & 20.3 & 41.6 & 44.5 & 41.2 & 8.6 & 29.0 & 18.4 & 39.8 & 34.6 & 29.4 \\
        \Xhline{5\arrayrulewidth}
    \end{tabular}
    }
    \caption{Average precision (AP), relative number of part evaluations (RNPE), and relative wall-clock time speedup (Speedup) of ADPM versus Cascade on all $20$ classes in PASCAL VOC 2007 and 2010.}
    \label{table:speedupCascade}
    
\end{table*}
%\newcolumntype{Y}{>{\centering\arraybackslash}X}

\begin{table*}
\centering
    \renewcommand{\arraystretch}{1.1}% Wider
    \resizebox{\textwidth}{!}{%
    \begin{tabular}{|l*{9}{|c}|}
    \Xhline{2\arrayrulewidth}
    & PCA no cache & PCA  cache & PE & Full no cache & Full cache & PE & Total no cache & Total cache & Total PE \\
    \Xhline{3\arrayrulewidth}
    CASCADE  &  4.34s & 0.67s & 208K     &   0.13s &   0.08s    &  1.1K  & 4.50s & 0.79s & 209K \\
    \hline
    ADPM  &       0.62s    &  0.06s & 36K     &   0.06s   &  0.04s     & 0.6K &  0.79s & 0.19s & 37K \\
    \Xhline{2\arrayrulewidth}
    \end{tabular}
    }
    \caption{An example demonstrating the computational time breakdown during inference of ADPM and Cascade on a single image. The number of part evaluations (PE) and the inference time (in seconds) is recorded for the PCA and the full-dimensional stages. The results are reported once without and once with cache use. The number of part evaluations is independent of caching. The total times are not equal to the sum of the two stages because of the additional but minimal time spent in I/O operations.}
    \label{table:timeanalysis}

\end{table*}%

%%%%%%%%%%%%%%%%%%%%%%%%%%%%%%%%% Cascade Multi threshold %%%%%%%%%%%%%%%%%%%%%%%%%
    % \centering
    % \renewcommand{\arraystretch}{1.2}% Wider
    % \resizebox{0.9\textwidth}{!}{%
    % \setlength{\tabcolsep}{.6em}
    % \begin{tabular}{r*{11}{c}}
    %     \Xhline{3\arrayrulewidth}
    %     Precision Level  &  0.0   & 0.1  & 0.2  & 0.3  & 0.4  & 0.5  & 0.6  & 0.7  & 0.8  & 0.9  & 1.0  \\
    %     \Xhline{3\arrayrulewidth}
    %     Cascade \#RNPE & 16.5 & 11.0 & 10.6 & 9.60 & 8.91 & 8.10 & 7.48 & 6.85 & 5.82 & 4.53 & 2.89 \\
    %     \Xhline{1\arrayrulewidth}
    %      Cascade mAP & 33.7 & 32.8 & 32.0 & 30.9 & 30.2 & 28.1 & 26.4 & 24.8 & 21.6 & 17.0 & 10.5 \\
    %      ADPM mAP  & 33.3 & - & - & - & - & - & - & - & - & - &  - \\ 
    %     \Xhline{3\arrayrulewidth}
    % \end{tabular}%
    % }
    % \caption{Mean average precision (mAP) and relative number of filter evaluations (\#RNPE) of Cascade versus ADPM on average as a function of threshold corresponding to precision levels, over all $20$ classes in PASCAL 2007.}
    % \label{table:apspeedup}
%%%%%%%%%%%%%%%%%%%%%%%%%%%%%%%%% Cascade Multi threshold %%%%%%%%%%%%%%%%%%%%%%%%%

%and INRIA pedestrian dataset\cite{DT05} 

In this section we compare ADPM\footnote{ADPM code and trained policies are available at:\\ \url{http://cis.upenn.edu/~menglong/adpm.html}} versus two baselines, the DPM and the cascade DPM (Cascade) in terms of average precision (AP), relative number of part evaluations (RNPE), and relative wall-clock time speedup (Speedup). Experiments were carried out on all 20 classes in the PASCAL VOC 2007 and 2010 datasets. Publicly available PASCAL 2007 and 2010 DPM and Cascade models were used for all three methods. 

\begin{figure*}[t!]
    \centering
    \raisebox{2pt}{\includegraphics[width=0.2\textwidth]{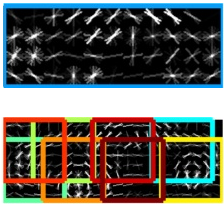}}
        \includegraphics[width=0.79\textwidth]{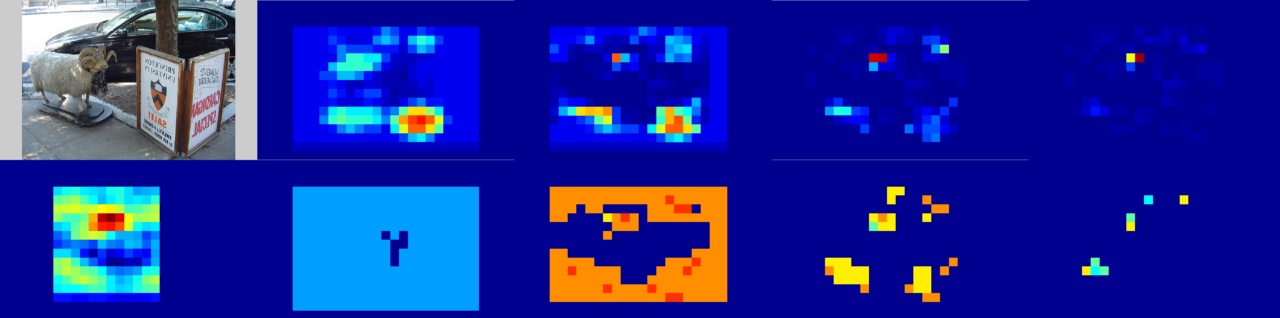}

    % \begin{subfigure}[b]{\textwidth}
    %     \raisebox{-2pt}{\includegraphics[width=0.2\textwidth]{figures/aero_model.png}}
    %     \includegraphics[width=0.79\textwidth]{figures/exp_aero.jpg}
    %     \caption{class: aeroplane}
    % \end{subfigure}
    % \begin{subfigure}[b]{\textwidth}
    %        \raisebox{23pt}{\includegraphics[width=0.2\textwidth]{figures/motor_model.png}}
    %        \includegraphics[width=0.79\textwidth]{figures/exp_motor.jpg}
    %        \caption{class: motorbike}
    %        \label{fig:exp_motor}
    % \end{subfigure}
    %\begin{subfigure}[b]{\textwidth}
    %    \raisebox{2pt}{\includegraphics[width=0.2\textwidth]{fig/car_model.png}}
    %    \includegraphics[width=0.79\textwidth]{fig/exp_car.jpg}
    %    \caption{class: car}
    %\end{subfigure}
    \caption{Illustration of the ADPM inference process on a car example. The DPM model with colored root and parts is shown on the left. The top row on the right consists of the input image and the evolution of the positive label probability ($p_t$) for $t \in \{1,2,3,4\}$ (high values are red; low values are blue). The bottom row consists of the full DPM $score(x)$ and a visualization of the parts applied at different locations at time $t$. The pixel colors correspond to the part colors on the left. In this example, despite the car being heavily occluded, ADPM converges to the correct location after four iterations.}

    \label{fig:exp}
\end{figure*}

\textbf{ADPM vs DPM}: The inference process of ADPM is shown in detail on two input images in Fig. \ref{fig:big_picture} and Fig. \ref{fig:exp}. The probability of a positive label $p_t$ (top row) becomes more contrasted as additional parts are evaluated. The number of locations at which the algorithm has not terminated decreases rapidly as the time progresses. Visually, the locations with a maximal posterior are identical to the top scores obtained by the DPM. The order of parts chosen by the policy is indicative of their informativeness. For example, in Fig. \ref{fig:exp} the wheel filters are applied first which agrees with intuition. In this example, the probability $p_t$ remains low at the correct location for several iterations due to the occlusions. Nevertheless, the policy recognizes that it should not terminate and as more parts are evaluated, the posterior reflects the correct location of the highest DPM score.

\begin{figure*}[t!]
    \centering
    \begin{subfigure}[b]{0.24\textwidth}
        \includegraphics[width=\textwidth]{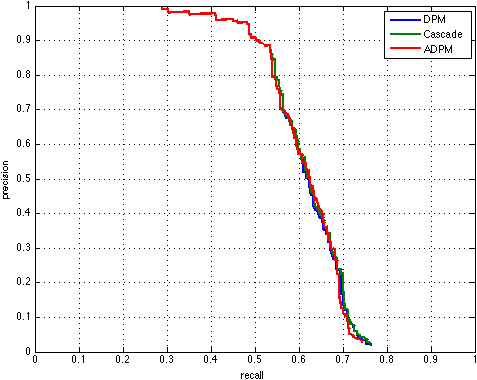}
        \caption{class: bicycle}
        \label{fig:bicycle}
    \end{subfigure}%
    %add desired spacing between images, e. g. ~, \quad, \qquad etc.
    %(or a blank line to force the subfigure onto a new line)
    \begin{subfigure}[b]{0.24\textwidth}
        \includegraphics[width=\textwidth]{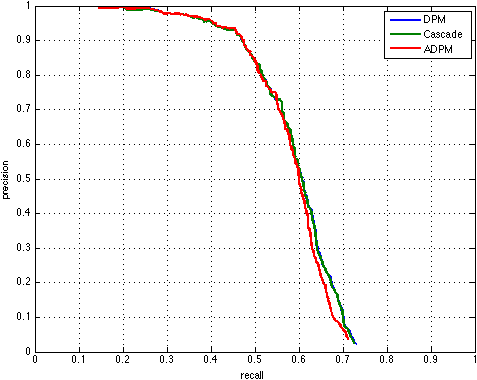}
        \caption{class: car}
        \label{fig:car}
    \end{subfigure}
    %add desired spacing between images, e. g. ~, \quad, \qquad etc.
    %(or a blank line to force the subfigure onto a new line)
    \begin{subfigure}[b]{0.24\textwidth}
        \includegraphics[width=\textwidth]{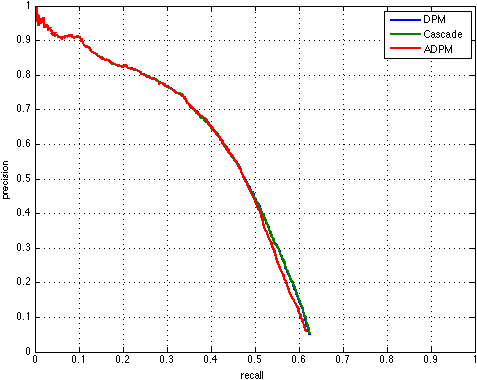}
        \caption{class: person}
        \label{fig:person}
    \end{subfigure}%
    \begin{subfigure}[b]{0.24\textwidth}
        \includegraphics[width=\textwidth]{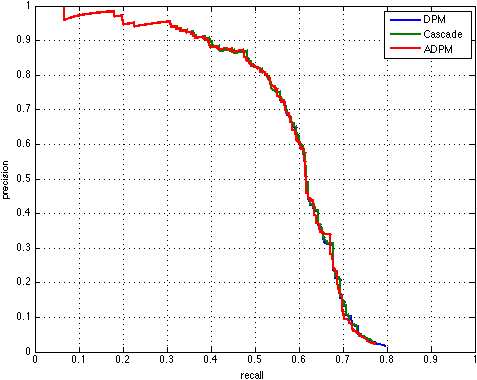}
        \caption{class: horse}
        \label{fig:horse}
    \end{subfigure}
    \caption{Precision recall curves for bicycle, car, person, and horse classes from PASCAL 2007. Our method's accuracy ties with the baselines.}
    \label{fig:pr}
\end{figure*}

ADPM was compared to DPM in terms of AP and RNPE to demonstrate the ability of ADPM to reduce the number of necessary part evaluations with minimal loss in accuracy irrespective of the features used. The ADPM parameters were set to $\lambda_{fp} = 20$ and $\lambda_{fn} = 5$ based on the analysis in Sec. \ref{sec:speed_acc}. Table \ref{table:speedupDPM} shows that ADPM achieves a significant decrease (about $90$ times on average) in the number of evaluated parts compared to DPM, while the loss in accuracy is negligible. The precision-recall curves of the two methods are shown for several classes in Fig. \ref{fig:pr}.
% with negligible loss in AP. Almost two orders of magnitude less part-locations are explored during inference compared to DPM. 

% Two sets of experiments were carried out. In the first, ADPM was compared to DPM in terms of AP and RNPE to demonstrate the ability of ADPM to reduce the number of necessary part evaluations with minimal loss in accuracy irrespective of the features used. In the second, ADPM was compared to Cascade in terms of AP, RNPE, and wall-clock time speedup to demonstrate the improvement in detection speed in seconds.

\textbf{ADPM vs Cascade}: The improvement in detection speed achieved by ADPM is demonstrated via a comparison to Cascade in terms of AP, RNPE, and wall-clock time (in seconds). Note that Cascade's implementation makes use of PCA-projected (top five dimensions) HOG features, which are very fast to compute. During inference, Cascade prunes the image locations in two passes. In the first pass, the locations are filtered using the PCA-projections and the low-scoring ones are discarded. In the second pass, the remaining locations are filtered using the full-dimensional features. To make a fair comparison, we adopted a similar two-stage approach for the active part selection. An additional policy was learned using PCA score likelihoods and was used to schedule PCA filters during the first pass. The locations, which were selected as foreground in the first stage, were filtered again, using the original policy to select the order of the full-dimensional filters. The parameters $\lambda_{fp}$ and $\lambda_{fn}$ were set to 20 and 5 for the PCA policy and to 50 and 5 for the full-dimensional policy. A higher $\lambda_{fp}$ was chosen to make the prediction more precise (albeit slower) during the second stage. Deformation pruning was not used for either method. Table \ref{table:speedupCascade} summarizes the results.

A discrepancy in the speedup of ADPM versus Cascade is observed in Table \ref{table:speedupCascade}. On average, ADPM is 7 times faster than Cascade in terms of RNPE but only 3 times faster in seconds. A breakdown of the computational time during inference on a single image is shown in Table \ref{table:timeanalysis}. We observe that the ratios of part evaluations and of seconds are consistent within individual stages (PCA and full). However, a single filter evaluation during the full-filter stage is significantly slower than one during the PCA stage. This does not affect the cumulative RNPE but lowers the combined seconds ratio. While ADPM is significanlty faster than Cascade during the PCA stage, the speedup (in seconds) is reduced during the slower full-dimensional stage.

% PR figure

%% file: tex/Conclusion.tex
\section{Conclusion}
This paper presents an active part selection approach which substantially speeds up inference with pictorial structures without sacrificing accuracy. Statistics learned from training data are used to pose an optimization problem, which balances the number of part filter convolution with the classification accuracy. Unlike existing approaches, which use a pre-specified part order and hard stopping thresholds, the resulting part scheduling policy selects the part order and the stopping criterion adaptively based on the filter responses obtained during inference. Potential future extensions include optimizing the part selection across scales and image positions and detecting multiple classes simultaneously.